\documentclass{article}
\usepackage{spconf,amsmath,graphicx}
\usepackage{booktabs}
\usepackage{verbatim}
\usepackage{color}
\usepackage{amsfonts}
\usepackage{multirow}
\usepackage{setspace}
\usepackage{amssymb}
\usepackage{pifont}
\usepackage{pgfplots}
\usepackage{tikz}
\usepackage{subfigure}
\usepackage{caption}
\usepackage{threeparttable}

\title{All you need is a second look: Towards Tighter Arbitrary shape text detection}
\name{Meng Cao$^{1}$ \qquad Yuexian Zou$^{1,2,*}$ 
\qquad \thanks{This paper was partially supported by National Engineering Laboratory for Video Technology - Shenzhen Division, Shenzhen Municipal Development and Reform Commission (Disciplinary Development Program for Data Science and Intelligent Computing).Special acknowledgements are given to Aoto-PKUSZ Joint Lab for its support.}}

\address{$^{1}$ ADSPLAB, School of ECE, Peking University, Shenzhen, China \\
$^{2}$Peng Cheng Laboratory, Shenzhen, China \\
*Corresponding author: zouyx@pku.edu.cn} 

\begin{document}
\bibliographystyle{unsrt} 
\topmargin=0mm
%
\maketitle
\begin{abstract}
Deep learning-based scene text detection methods have progressed substantially over the past years. However, there remain several problems to be solved. Generally, long curve text instances tend to be fragmented because of the limited receptive field size of CNN. Besides, simple representations using rectangle or quadrangle bounding boxes fall short when dealing with more challenging arbitrary-shaped texts. In addition, the scale of text instances varies greatly which leads to the difficulty of accurate prediction through a single segmentation network. To address these problems, we innovatively propose a two-stage segmentation based arbitrary text detector named \textit{NASK} (\textbf{N}eed \textbf{A} \textbf{S}econd loo\textbf{K}). 
Specifically, \textit{NASK} consists of a Text Instance Segmentation network namely \textit{TIS} (\(1^{st}\) stage), a Text RoI Pooling module and a Fiducial pOint eXpression module termed as \textit{FOX} (\(2^{nd}\) stage). 
Firstly, \textit{TIS} conducts instance segmentation to obtain rectangle text proposals with a proposed Group Spatial and Channel Attention module (\textit{GSCA}) to augment the feature expression.
Then, Text RoI Pooling transforms these rectangles to the fixed size. Finally, \textit{FOX} is introduced to reconstruct text instances with a more tighter representation using the predicted geometrical attributes including text center line, text line orientation, character scale and character orientation. Experimental results on two public benchmarks including \textit{Total-Text} and \textit{SCUT-CTW1500} have demonstrated that the proposed \textit{NASK} achieves state-of-the-art results.



\end{abstract}
\section{Introduction}
\label{sec:intro}

Recently, scene text detection (STD) in the wild has drawn extensive attention because of its practical applications\cite{zhu2017cascaded}, such as  blind navigation, autonomous driving, \textit{etc}. Generally, the performance of STD has been greatly enhanced by the advanced object detection\cite{girshick2015fast}\cite{liu2016ssd} and segmentation\cite{long2015fully} frameworks which can be divided into two categories: 1) Segmentation-based methods\cite{zhang2016multi}\cite{yao2016scene}. These methods draw inspiration from instance segmentation and conduct dense predictions in pixel levels. 2) Regression-based methods\cite{liao2018textboxes++}\cite{zhou2017east}. Scene texts are detected using the adapted one-stage or two-stage frameworks which have been proved effective in general object detection tasks.


However, STD remains a challenging task due to its unique characteristics. Firstly, since the convolutional operation which is widely used in all segmentation or detection networks only processes a \textit{local} neighbour, it hardly captures the long-range dependencies even it is stacked repeatedly. Thus, CNN-based STD methods sometimes fail to detect long text instances because they are far beyond CNN's receptive field\cite{wang2018non}. Secondly, although detecting words or text lines with a relatively simple rectangle or quadrilateral representation has been well tackled,  curve text detection with a more tight representation is not well solved\cite{long2018textsnake}. 
Finally, some text instances are extremely tiny which makes their precise shape description more difficult because even a little segmentation deviation may lead to the ultimate failure. Therefore, a single segmentation network fails to process images that vary greatly in text scales.



In order to solve the problems mentioned above, we propose \textit{NASK} which contains a Text Instance Segmentation network (\textit{TIS}) and a Fiducial pOint eXpression module (\textit{FOX}), connected by Text RoI pooling. \textit{TIS} is a context attended FCN\cite{long2015fully} with a proposed Group Spatial and Channel Attention (\textit{GSCA}) for text instance segmentation. \textit{GSCA} captures long-range dependencies by directly computing interactions between any two positions across both space and channels, which enhances the semantic information of shared feature maps. Then, similar to Faster R-CNN\cite{ren2015faster}, Text RoI pooling accepts the shared feature maps and the bounding box coordinates generated by \textit{TIS} as input and "warps" these rectangular RoIs into a fixed size. Finally, \textit{FOX} reconstructs texts with a set of fiducial points which are calculated using the predicted geometry attributes.

The main contributions of this work are summarized as follows: (a) A group spatial and channel attention module (\textit{GSCA}) which aggregates the contextual information is introduced into FCN for feature refinements. (2) We propose a Fiducial pOint eXpression module (\textit{FOX}) for the tighter arbitrary shape text detection. (3) A novel two-stage segmentation based STD detector named \textit{NASK} incorporating \textit{GSCA} and \textit{FOX} is trained jointly and achieves state-of-the-art performance on two curve text detection benchmarks. 




\begin{figure}[htb]
 \centering
  \centerline{\includegraphics[width=9cm]{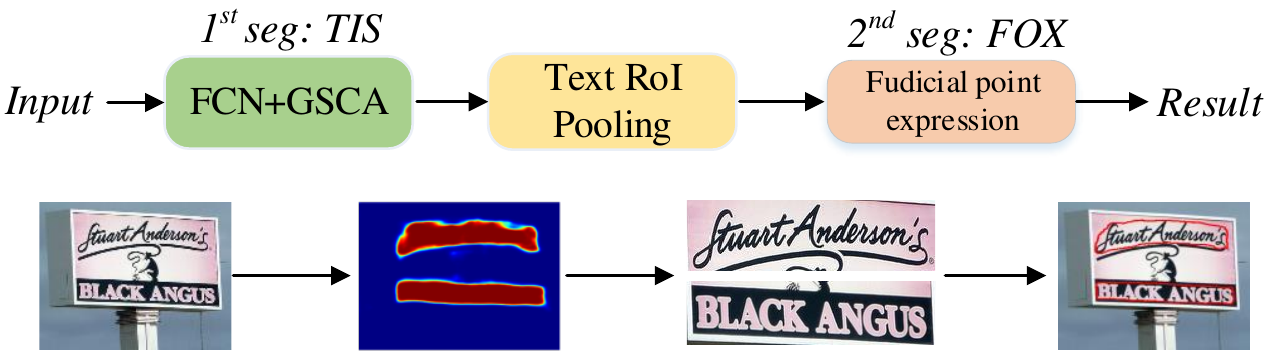}}
\caption{The pipeline of NASK. \textbf{Above:} \(1^{st}seg\) and \(2^{nd}seg\) means the first and the second stage segmentation network respectively. \textbf{Below:} The illustration of an image instance transformation process.}
\label{fig:res}
\end{figure}

\section{Approach}
\label{sec:format}
In this section, we describe the pipeline of the proposed \textit{NASK}. Firstly, the overall pipeline of the whole model is briefly described in Section 2.1. Next, we elaborate on all proposed modules including \textit{GSCA} and \textit{FOX}. Finally, the optimization details are given in Section 2.4.   

\subsection{Overview}
The overall architecture is demonstrated in Fig 1. Firstly, ResNet-50 based FCN with \textit{GSCA} makes up the first stage text instance segmentation network \textit{TIS}.
Then Text RoI Pooling module transforms the rectangle text proposals to a fixed size. Finally, \textit{FOX} is applied to obtain a tighter representation of curve text instances.

\subsection{Group Spatial and Channel Attention Module }


Inspired by Non-local network\cite{wang2018non} which is based on the self-attention mechanism\cite{vaswani2017attention}, a Group Spatial and Channel Attention module is proposed. The detailed structure is displayed in Fig 2. Compared to Non-local network which only models the interactions between spatial positions in the same channel, \textit{GSCA} explicitly learns the correlations among all elements across both space and channels. In order to alleviate the huge computational burden, \textit{GSCA} incorporates the channel grouping idea to gather all \textit{C} channels into \textit{G} groups. Only the relationships within each group which contains \(C' = C/G\) channels are calculated and the computational complexity decreases from \((H\times W\times C)^2\) to \(G\times(H\times W\times C/G)^2 = (H\times W\times C)^2/G\). As for the affiliation among different groups, similar to SENet\cite{hu2018squeeze}, the branch of global channel attention in Fig 2 is set to generate global channel-wise attention and distribute information among every group.

Specifically, the attended feature map is expressed as \(Y = f(\Theta(X), \Phi(X))g(X)\). Here \(\Theta(X), \Phi(X)\) are learnable spatial transformations implemented as serially connected \textit{convolution} and \textit{reshape} while \(f(\cdot,\cdot)\) is defined as matrix product for simplification. Then we have \(Y' = \Theta(X)\Phi(X)^Tg(X)\) where \(Y'\) is the \textit{group} result of \(Y\). Another branch aiming to capture global channel weights is implemented with two convolution layers and one fully connected layer. Thus, through \(Weighted Concat\), we deduce \(Y = concat(\lambda_i Y'_i)\), \(i=0,1,...,C-1,\) where $C$, $\lambda_i$ and $Y'_i$ denote the number of channels, $i$-$th$ channel weight and $i$-$th$ channel feature map respectively. Meanwhile, a short-cut path is used to preserve the local information and the final output can be written as \(Z = X + Y\).

\begin{figure}[htb]
 \centering
  \centerline{\includegraphics[width=9cm]{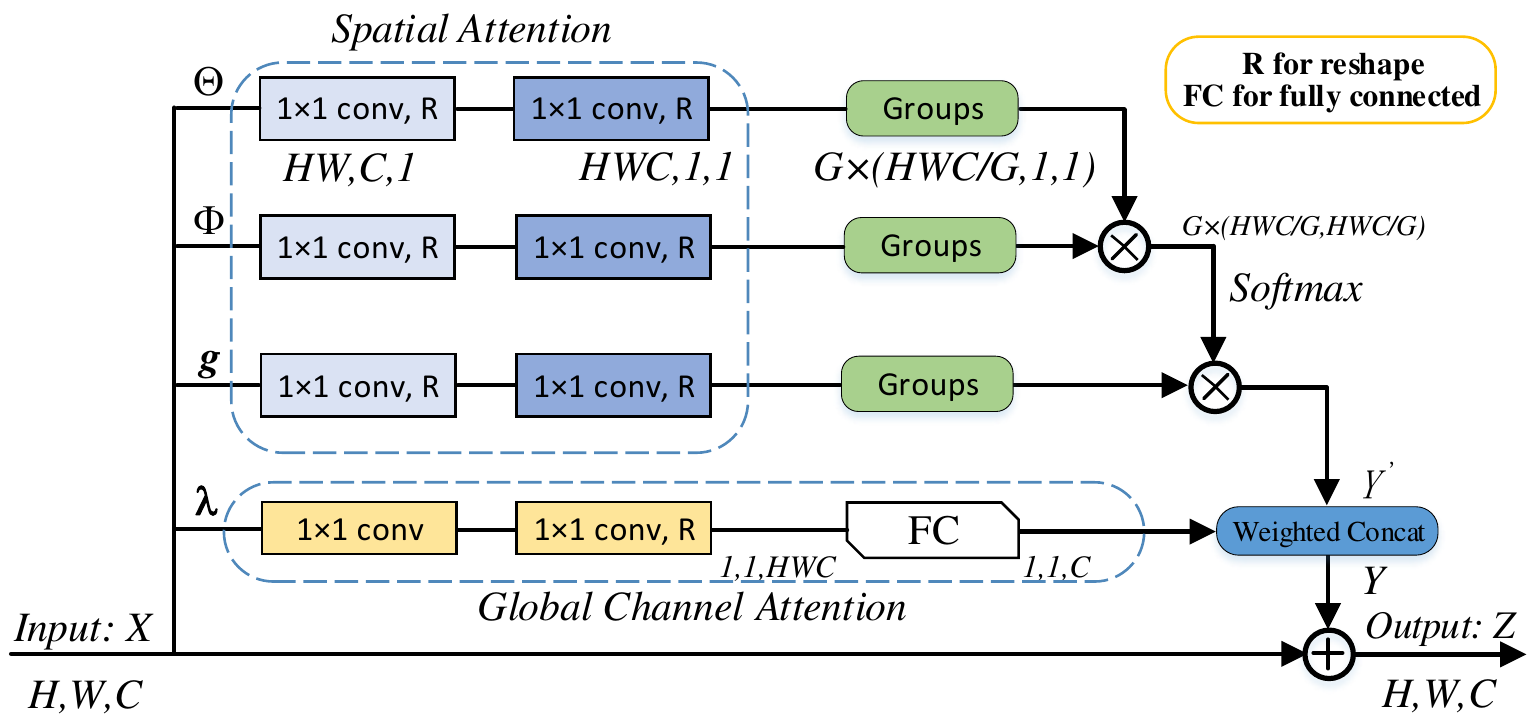}}
\caption{Group Spatial and Channel Attention module: Intra-group attention is learned by the serially connected spatial \textit{convolution} and \textit{reshape} denoted as \textit{\(\Theta\)}, \(\Phi\) while the global channel attention is captured by transformation \(\lambda\). "\(\oplus\)" denotes the element-wise sum while "\(\otimes\)" denotes matrix multiplication. The annotation under each block represents the corresponding output size.}
\label{fig:res}
\end{figure}


\subsection{Fiducial Point Expression Module}

\begin{figure}[htb]
 \centering
  \centerline{\includegraphics[width=8cm]{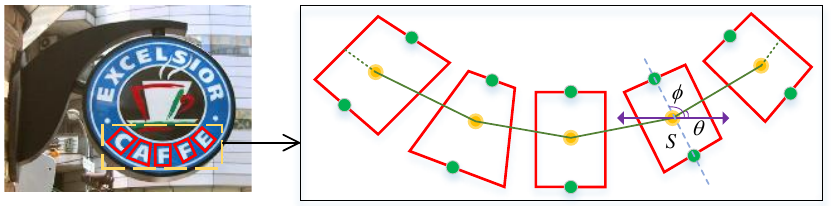}}
\caption{Illustration of Fiducial points expression module.}
\label{fig:res}
\end{figure}

As depicted in Fig 3, the geometrical representation of text instances includes text center line (\textit{TCL})  , character scale \(s\), character orientation \( \phi \) and text orientation \( \theta \). Specifically, the text center line is a binary mask based on the side-shrunk version of text polygon annotations. The scale \(s_i\) is half the height of the character while the text orientation \( \theta_i\) is defined as the horizontal angle between the current quadrilateral center \(c_i\) and the next  one \(c_{i+1}\). We take the midpoints on the top and bottom edges of each character quadrilateral as fiducial points and the character orientation \( \phi_i\) is defined as the direction from the midpoint of the bottom edge to that of the top edge.



Mathematically, a text instance can be viewed as an ordered sequence \(S = \{S_1, ..., S_i,...,S_n\}\), where \(n\) is a hyper-parameter which denotes the number of character segments. Each node \(S_i\) is associated with a group of geometrical attributes and can be represented as \(S_i = (c_i,s_i,\phi_i, \theta_i)\) 
where every element is defined as above.



\begin{figure}[htb]
 \centering
    \centerline{\includegraphics[width=8.5cm]{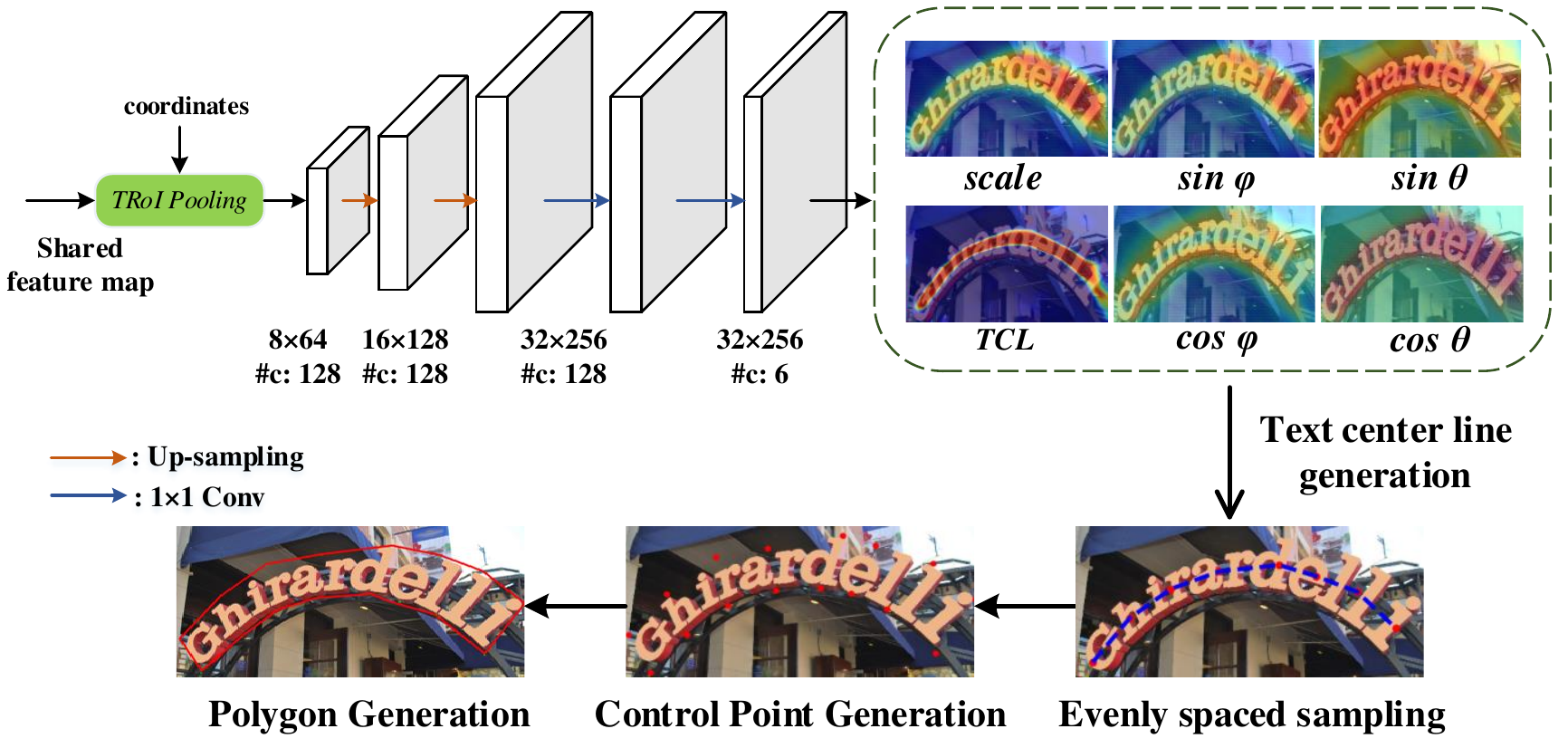}}
\caption{The illustration of Fiducial Point Expression. Note that, \#c stands for the number of channels.}
\label{fig:res}
\vspace{-6pt}
\end{figure}

The overall text polygon generation process is illustrated in Fig 4. Firstly, two up-sampling and one \(1\times1\) convolution with 6 output channels are applied to regress all the geometrical attributes. The output is \(F=\{f_1,f_2,...,f_6\}\) where $f_1$,$f_2$ denote the character scale \(s\) of each pixel and the probability of pixels on \textit{TCL} respectively. \(sin \theta\) and \(cos \theta\) are normalized as \(cos \theta = \frac{f_3}{\sqrt{f_3^2+f_4^2}}\), \(sin \theta = \frac{f_4}{\sqrt{f_3^2+f_4^2}}\) to ensure their quadratic sum equals to 1. $sin \phi$ and $cos \phi$ are normalized in the same way. Then \(n\) points are equidistantly sampled in the center line \(C\), named \( \overline{C} = ( \overline{c}_1, ..., \overline{c}_i, ..., \overline{c}_n ) \). For each \(\overline{c}_i\), according to the geometric relationship, two corresponding fiducial points are computed as follows.
\begin{equation} \label{eqn2}
  \begin{split}
   p_{2i-1} &= \overline{c}_i + (s_i cos\phi_i, -s_i sin\phi_i),    \\
   p_{2i} &= \overline{c}_i + (-s_i cos\phi_i, s_i sin\phi_i).
  \end{split}
\end{equation}
where \(\overline{c}_i\), \(s_i\), \(\phi_i\) are the center coordinate, scale and orientation for the \(i\)-\(th\) character respectively. Therefore, one single text instance can be represented with \(2n\) fiducial points. Finally, text polygons are generated by simply applying \textit{approxPolyDP} in \textit{OpenCV}\cite{bradski2008learning} and then mapped back to the original image proportionally.

\subsection{Optimization}
The whole network is trained in an end-to-end manner using the following loss function:

\begin{equation}
    L = \lambda_0L_{TIS} + L_{FOX}
\end{equation}
where \(L_{TIS}\) and \(L_{FOX}\) are the loss for Text Instance Segmentation and Fiducial Point Expression module respectively. \(L_{TIS}\) is cross-entropy loss for text regions with OHEM\cite{shrivastava2016training} adopted. 
For \(L_{FOX}\), it can be expressed as follows:
\begin{equation}
  \begin{split}
    L_{FOX} = &\lambda_1 L_{tcl} + \lambda_2 L_{s} + \lambda_3 L_{sin\theta} \\
    &+ \lambda_4 L_{cos \theta} + \lambda_5 L_{sin\phi} + \lambda_6 L_{cos\phi}
    \end{split}
\end{equation}
where \(L_{tcl}\) is cross-entropy loss for \textit{TCL}. \(L_s\),\(L_{sin\theta}\),\(L_{cos\theta}\),\(L_{sin\phi}\) and \(L_{cos\phi}\) are all calculated using Smoothed-L1 loss. All pixels outside \textit{TCL} are set to 0 since the geometrical attributes make no sense to \textit{non-TCL} points. The hyper-parameters \(\lambda_0,\lambda_1,\lambda_2,\lambda_3,\lambda_4,\lambda_5,\lambda_6\) are all set to 1 in our experiments.

\section{Experiments}
\label{sec:pagestyle}
To evaluate the effectiveness of the proposed \textit{NASK}, we adopt two widely used datasets with arbitrary shape text instances for experiments and present detailed ablation studies.

\subsection{Datasets}

\textbf{Total-Text\cite{ch2017total}} is a newly-released dataset for curve text detection which contains horizontal and multi-oriented texts as well. It is split into training and testing sets with 1255 and 300 images respectively.

\textbf{SCUT-CTW1500\cite{yuliang2017detecting}} is a challenging dataset for long curve text detection. It consists of 1000 training images and 500 testing images. The text instances from this dataset are annotated as polygons with 14 vertices. 



\subsection{Implementation Details}
The proposed method is implemented in PyTorch. For all datasets, images are randomly cropped and resized into \(512\times512\). The cropped image regions are rotated randomly in 4 directions with \(0^\circ\), \(90^\circ\), \(180^\circ\), \(270^\circ\).
The experiments are conducted on four NVIDIA TitanX GPUs each with 12GB memory. The training process is divided into two stages. Firstly, \(1^{st}\) stage segmentation network is trained using Synthetic dataset\cite{gupta2016synthetic} for 10 epochs. We take this step as a warm-up training strategy because the precise first-stage segmentation is a prerequisite for the subsequent text shape refinement. 
Then in the fine-tuning step, the whole model is trained using Adam optimizer with the learning rate re-initiated to \(10^{-4}\) and the learning rate decay factor set to 0.9. 

\subsection{Evaluation on Curved Text Benchmark}
We evaluate the performance of \textit{NASK} on Total-Text and SCUT-CTW1500 after finetuning about 10 epochs. The number of sample points \(n\) in \textit{TCL} is set to 8 and the group number \(G\) of GSCA is set to 4. Thresholds \(T_{tr}\), \(T_{tcl}\) for regarding pixels to be text regions or \(TCL\) are set to (0.7,0.6) and (0.8,0.4) respectively for Total-Text and SCUT-CTW1500. All quantitative results are shown in Table 1.

\begin{table}[!ht]
    \caption{Results of Total-Text and SCUT-CTW 1500}\label{tab:tablenotes}
    \centering
    \begin{threeparttable}
\setlength{\tabcolsep}{3pt}    
\begin{tabular}{|c|c|c|c|c|c|c|c|c|}
\hline
\multirow{2}*{Model} & \multicolumn{4}{c|}{Total-Text} & \multicolumn{4}{c|}{SCUT-CTW 1500 }\\
\cline{2-9}
~                     & \textit{R} & \textit{P} & \textit{H}  & \textit{F} & \textit{R} & \textit{P} & \textit{H}  & \textit{F}\\
\hline
Deconv\cite{ch2017total} & 33.0  & 40.0 &  36.0 & -  & - & - & - & -\\
TextField\cite{xu2019textfield} & 79.9 & 81.2 & 80.6 & - & -  & - & - & - \\
CTPN\cite{tian2016detecting} & -  & - & - & - & 53.8 & 60.4 & 56.9 & 7.14 \\
CTD\cite{yuliang2017detecting} & -  & - & - & - & 69.8 & 74.3 & 73.4 & 13.3 \\
SLPR\cite{zhu2018sliding} & -  & - & - & - &  70.1 & 80.1 & 74.8 & - \\
SegLink\cite{shi2017detecting} & 23.8 & 30.3  & 26.7 & - & 40.0 & 42.3 & 40.8 & 10.7 \\
EAST\cite{zhou2017east} & 36.2 & 50.0  & 42.0 & -  & 49.1 & 78.7 & 60.4 & \color{red}21.2 \\
PSENet\cite{li2018shape} & 75.1 & 81.8  & 78.3 & 3.9 & 75.6 & 80.6 & 78.0 & 3.9 \\
TextSnake\cite{long2018textsnake} & 74.5 & 82.7 & 78.4 & - & \color{red}85.3 & 67.9 & 75.6 & - \\
NASK(Ours) & \color{red}81.2 &  \color{red}83.3 &  \color{red}82.2  & \color{red}8.4  & 78.3 & \color{red}82.8 & \color{red}80.5 & 12.1 \\ 
\hline
\end{tabular} 
      \begin{tablenotes}
        \footnotesize
        \item Note: \textit{R},\textit{P},\textit{H},\textit{F} denotes Recall, Precision, H-mean and FPS respectively. For fair comparison, no external data is used for all models.
      \end{tablenotes}
    \end{threeparttable}
    \vspace{-5pt}
  \end{table}

From Table1, we can see that \textit{NASK} achieves the highest \textit{H-mean} value of 82.2\% with \textit{FPS} reaching 8.4 on Total-Text. The quantitative results on SCUT-CTW1500 dataset also show \textit{NASK} achieves a competitive result comparable to state-of-the-art methods with \textit{H-mean} and \textit{Precision} attaining 80.5\% and 82.8\%. Selected detection results are shown in Fig 5.


\begin{figure}[htbp]
    \centering
    \subfigure[Total Text]{
       \includegraphics[width=0.75in,height=0.7in]{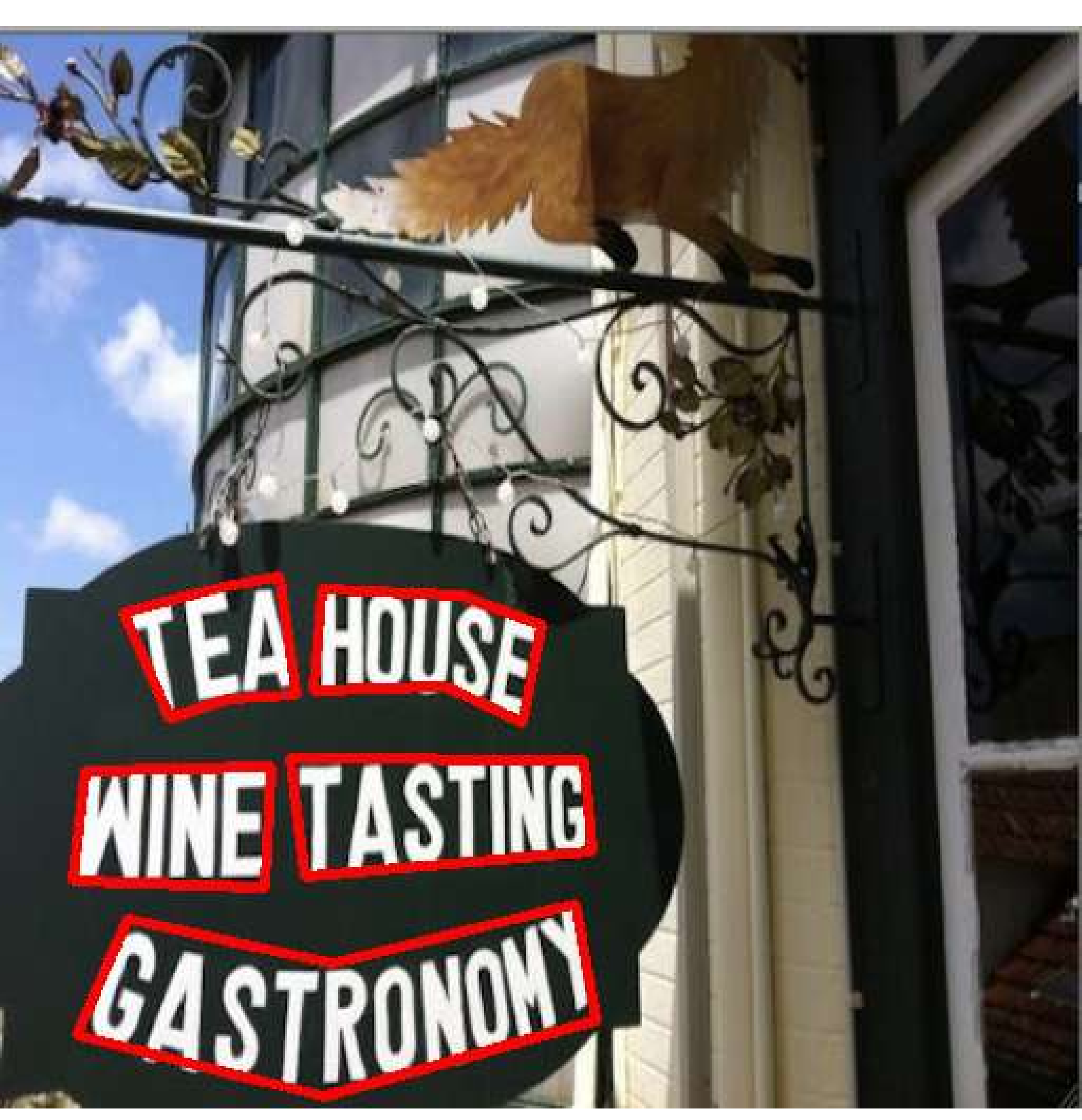}
       \includegraphics[width=0.75in,height=0.7in]{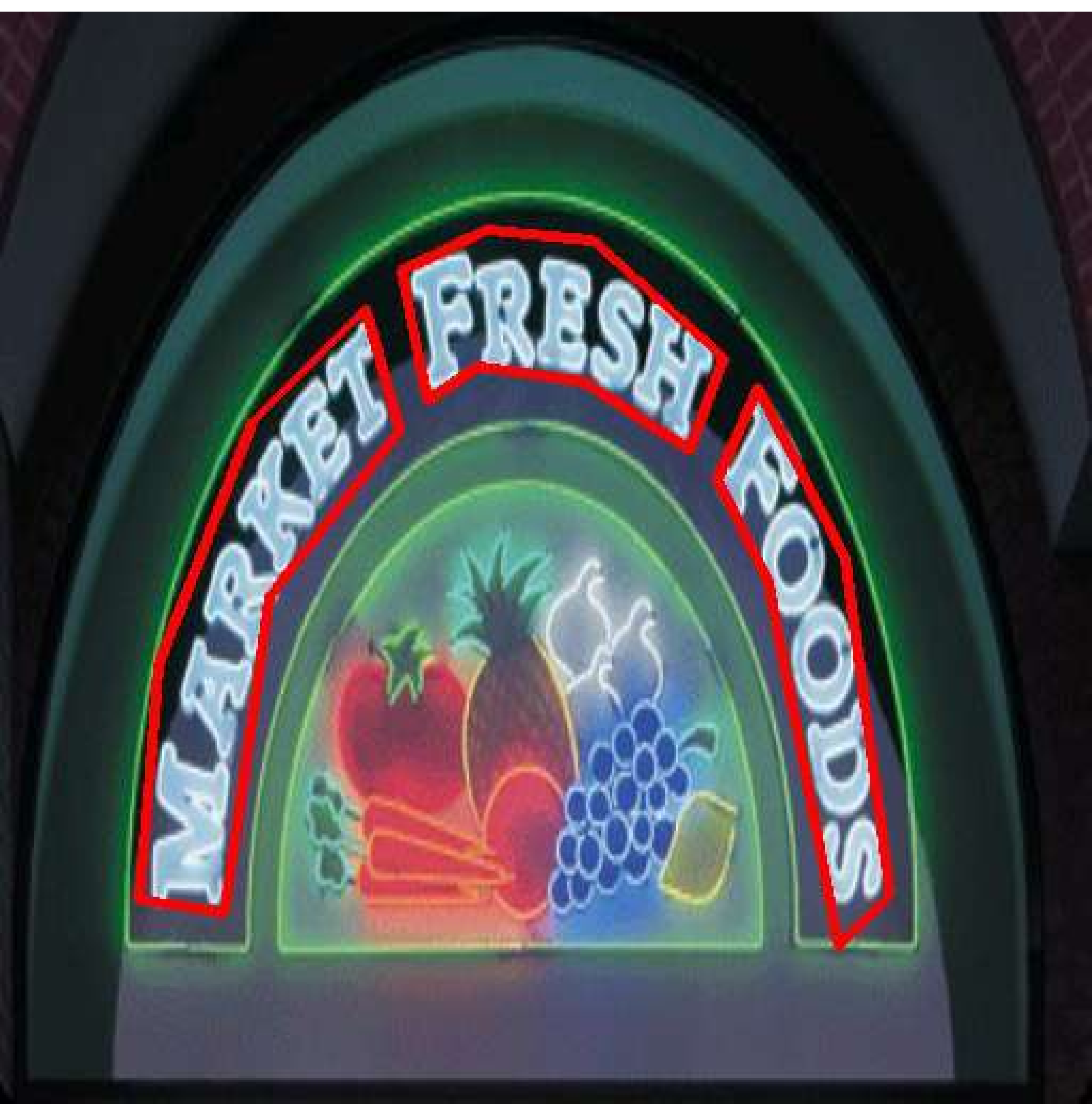}
    }
    \subfigure[SCUT-CTW 1500]{
       \includegraphics[width=0.75in,height=0.7in]{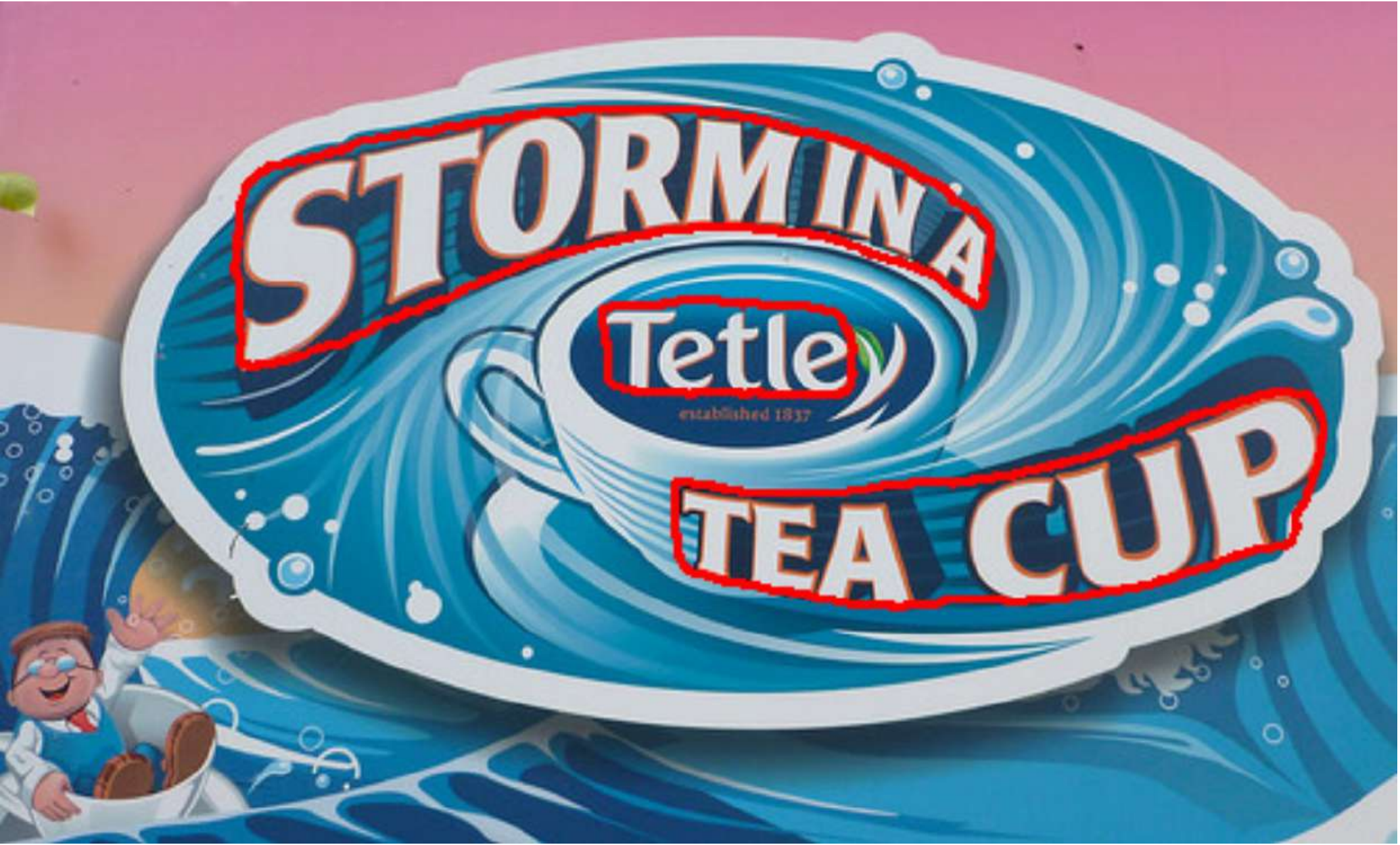}
       \includegraphics[width=0.75in,height=0.7in]{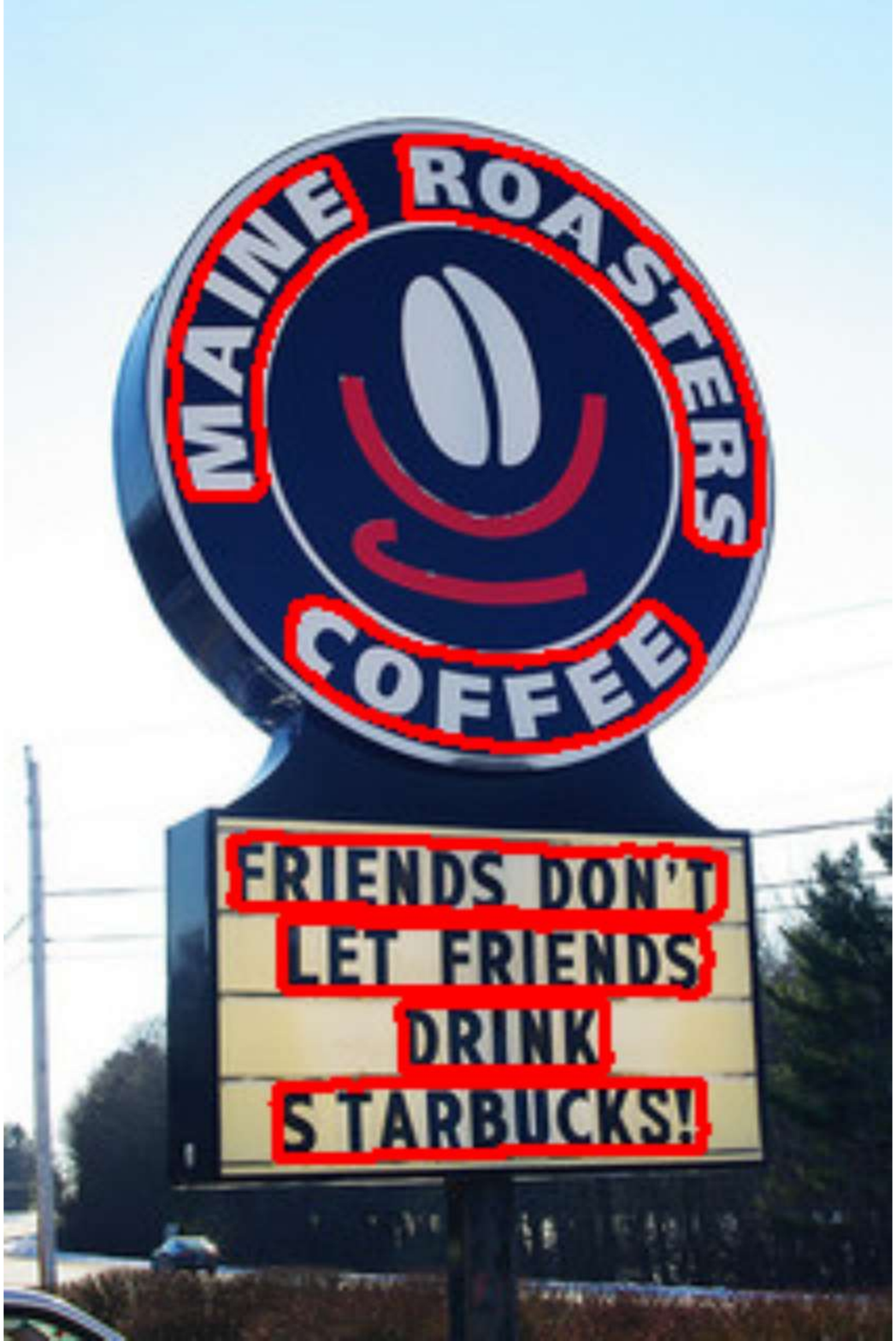}
    }
    \caption{Qualified detection results of Total Text and SCUT-CTW 1500.}
    \label{fig:my_label}
    \vspace{-4pt}
\end{figure}

\vspace{-4pt}
\subsection{Ablation studies}
We conduct several ablation experiments on SCUT-CTW1500 to analyze the proposed \textit{NASK}. Details are discussed as follows.

\textbf{Effectiveness of GSCA.}
We devise a set of comparative experiments to demonstrate the effectiveness of \textit{GSCA}. For fair comparisons, we replace \textit{GSCA} with two stacked \(1\times1\) convolution layers so that they share almost the same computation overhead. The experiment results in Table 2(a) show that \textit{GSCA} brings about an obvious ascent in performance. For instance, by setting \(G\) to 4, \textit{H-mean} improves by 2.5\% compared to the native model (\(G=0\)).
The visualization analysis in Fig 6 indicates that \textit{GSCA} is context-aware that most of the weights are focused on the pixels belonging to the same category with the \textit{reference} \textit{pixel}.

\begin{figure}[htb]
 \centering
    \centerline{\includegraphics[width=8cm]{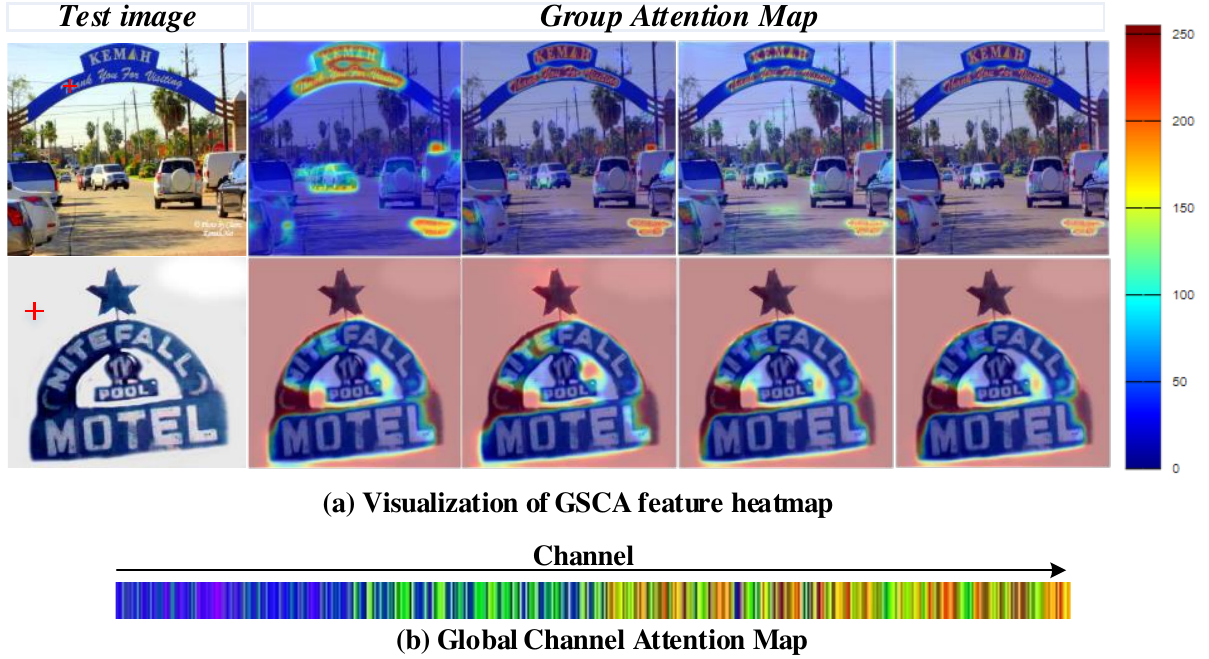}}
\caption{(a) Column 1: one image with a red cross marked \textit{reference pixel} which is a selected position in \(g(X)\) described in \textit{Sec} 2.2. Column 2 to 5: related feature heatmaps computed with GSCA. Specifically, we use the corresponding vectors in \(\Phi(X)\) and \(\Theta(X)\) to compute attention maps. (b) Global Channel Attention Map displays the weight distribution along channels.}
\label{fig:res}
\vspace{-6pt}
\end{figure}   

\begin{table}[!ht]
    \caption{Ablation studies on SCUT-CTW 1500}\label{tab:tablenotes}
    \centering
    \begin{threeparttable}
     
\begin{tabular}{|c|c|c|c|c|c|c|}
\hline
index & \(1^{st} seg\) & \textit{\(G\)} & \textit{R} & \textit{P} & \textit{H} & \textit{F} \\
\hline
\multirow{6}*{(a)} & \multirow{6}*{\ding{51}} & 0 & 76.4 &  79.7 & 78.0 & 16.3 \\
~ & ~ & 2 & 79.3 & 83.2 & 81.2 & 3.4 \\
~ & ~ & 4 & 78.3 & 82.8 & 80.5 & 12.1 \\
~ & ~ & 8 & 78.1 & 82.3 & 80.1 & 12.5 \\
~ & ~ & 12 & 77.9 & 82.8 & 80.3 & 12.7 \\
~ & ~ & 16 & 77.3 & 81.6 & 79.4 & 12.9 \\
\hline
\multirow{2}*{(b)} & \ding{55} & 4 & 72.7 & 77.5 & 75.0 & 16.3 \\
~ & \ding{51} & 4 & 78.3 & 82.8 & 80.5 & 12.1 \\
\hline 
\end{tabular} 
      \begin{tablenotes}
        \footnotesize
        \item Note: \(1^{st} seg\) means the first stage segmentation namely \textit{TIS}; \(G\) denotes the group number of the attention module. 
      \end{tablenotes}
    \end{threeparttable}
    \vspace{-5pt}
  \end{table}

\textbf{Influence of the number of attention module groups \textbf{\textit{G}}.}
Several experiments are operated to study the impact of the group number of \textit{GSCA} and the results are shown in Table 2(a). As expected, the detection speed increases with the rise of the group number and reaches the limit at about 12.9 \textit{FPS}. It is also worthwhile to notice that the detection result is not much sensitive to \(G\). This may be attributed to the fact that the global channel attention effectively captures the rich correlations among groups. 




\textbf{Influence of the number of sample points \textbf{\textit{n}}.}
The curve text representation is decided by a set of \(2n\) fiducial points. 
We evaluate \textit{NASK
} with different values of \(n\) and results are shown in Fig 7. The performance witnesses a gigantic increase when \(n\) changes from 2 to 8 and then gradually converges. Therefore, we set \(n\) to 8 in our experiments.

\begin{figure}[htb]
 \centering
  \centerline{\includegraphics[width=5cm]{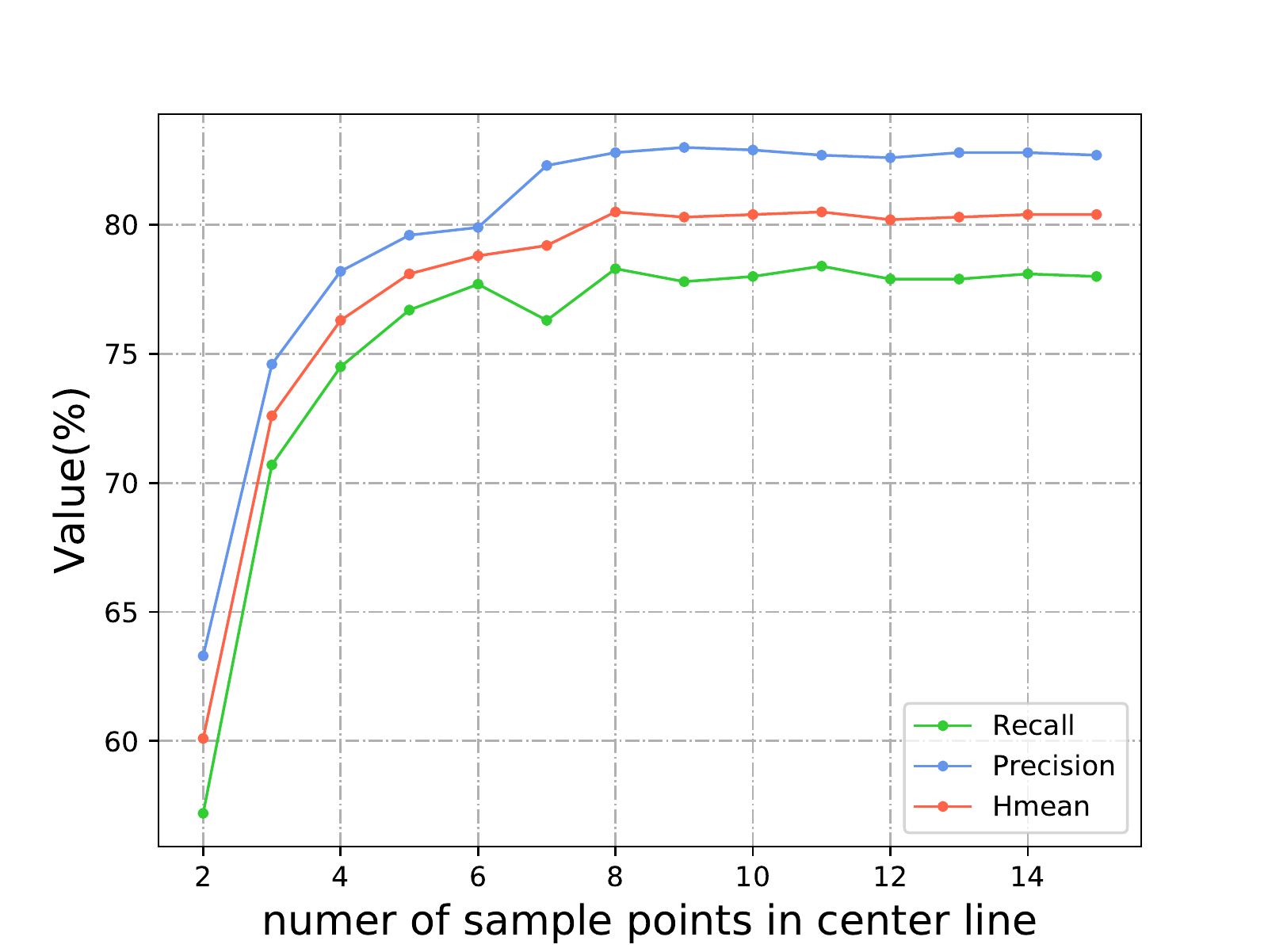}}
\caption{Ablations for the number of sample points.}
\label{fig:res}
\vspace{-6pt}
\end{figure}

\textbf{Effectiveness of the first-stage segmentation (\textit{TIS}).}
To demonstrate the effectiveness of the two-stage architecture, we conduct experiments that directly apply \textit{FOX} on the input image and the comparative results are list in Table 2(b). It is obvious that the two-stage segmentation network effectively improves the detection performance with \textit{H-mean} improved by 5.5\%.

\section{Conclusion}
In this paper, we propose a novel text detector \textit{NASK} to facilitate the detection of arbitrary shape texts. The whole network consists of serially connected Text Instance Segmentation (\textit{TIS}), Text RoI Pooling and Fiducial Point Expression module (\textit{FOX}). \textit{TIS} conducts text instance segmentation while Text RoI Pooling transforms rectangle text bounding boxes to the fixed size. Then \textit{FOX} achieves a tight and precise text detection result by predicting several geometric attributes. To capture the long-range dependency, a self-attention based mechanism called Group Spatial and Channel Attention module (GSCA) is incorporated into \textit{TIS} to augment the feature representation. The effectiveness and efficiency of the proposed \textit{NASK} have been proved by experiments with \textit{H-mean} reaching 82.2\% and 80.5\% for Total-Text and SCUT-CTW 1500 respectively.

\bibliography{egbib.bib}

\begin{thebibliography}{10}

\bibitem{zhu2017cascaded}
Yingying Zhu, Minghui Liao, Mingkun Yang, and Wenyu Liu.
\newblock Cascaded segmentation-detection networks for text-based traffic sign
  detection.
\newblock {\em IEEE transactions on intelligent transportation systems},
  19(1):209--219, 2017.

\bibitem{girshick2015fast}
Ross Girshick.
\newblock Fast r-cnn.
\newblock In {\em Proceedings of the IEEE international conference on computer
  vision}, pages 1440--1448, 2015.

\bibitem{liu2016ssd}
Wei Liu, Dragomir Anguelov, Dumitru Erhan, Christian Szegedy, Scott Reed,
  Cheng-Yang Fu, and Alexander~C Berg.
\newblock Ssd: Single shot multibox detector.
\newblock In {\em European conference on computer vision}, pages 21--37.
  Springer, 2016.

\bibitem{long2015fully}
Jonathan Long, Evan Shelhamer, and Trevor Darrell.
\newblock Fully convolutional networks for semantic segmentation.
\newblock In {\em Proceedings of the IEEE conference on computer vision and
  pattern recognition}, pages 3431--3440, 2015.

\bibitem{zhang2016multi}
Zheng Zhang, Chengquan Zhang, Wei Shen, Cong Yao, Wenyu Liu, and Xiang Bai.
\newblock Multi-oriented text detection with fully convolutional networks.
\newblock In {\em Proceedings of the IEEE Conference on Computer Vision and
  Pattern Recognition}, pages 4159--4167, 2016.

\bibitem{yao2016scene}
Cong Yao, Xiang Bai, Nong Sang, Xinyu Zhou, Shuchang Zhou, and Zhimin Cao.
\newblock Scene text detection via holistic, multi-channel prediction.
\newblock {\em arXiv preprint arXiv:1606.09002}, 2016.

\bibitem{liao2018textboxes++}
Minghui Liao, Baoguang Shi, and Xiang Bai.
\newblock Textboxes++: A single-shot oriented scene text detector.
\newblock {\em IEEE transactions on image processing}, 27(8):3676--3690, 2018.

\bibitem{zhou2017east}
Xinyu Zhou, Cong Yao, He~Wen, Yuzhi Wang, Shuchang Zhou, Weiran He, and Jiajun
  Liang.
\newblock East: an efficient and accurate scene text detector.
\newblock In {\em Proceedings of the IEEE conference on Computer Vision and
  Pattern Recognition}, pages 5551--5560, 2017.

\bibitem{wang2018non}
Xiaolong Wang, Ross Girshick, Abhinav Gupta, and Kaiming He.
\newblock Non-local neural networks.
\newblock In {\em Proceedings of the IEEE Conference on Computer Vision and
  Pattern Recognition}, pages 7794--7803, 2018.

\bibitem{long2018textsnake}
Shangbang Long, Jiaqiang Ruan, Wenjie Zhang, Xin He, Wenhao Wu, and Cong Yao.
\newblock Textsnake: A flexible representation for detecting text of arbitrary
  shapes.
\newblock In {\em Proceedings of the European Conference on Computer Vision
  (ECCV)}, pages 20--36, 2018.

\bibitem{ren2015faster}
Shaoqing Ren, Kaiming He, Ross Girshick, and Jian Sun.
\newblock Faster r-cnn: Towards real-time object detection with region proposal
  networks.
\newblock In {\em Advances in neural information processing systems}, pages
  91--99, 2015.

\bibitem{vaswani2017attention}
Ashish Vaswani, Noam Shazeer, Niki Parmar, Jakob Uszkoreit, Llion Jones,
  Aidan~N Gomez, {\L}ukasz Kaiser, and Illia Polosukhin.
\newblock Attention is all you need.
\newblock In {\em Advances in neural information processing systems}, pages
  5998--6008, 2017.

\bibitem{hu2018squeeze}
Jie Hu, Li~Shen, and Gang Sun.
\newblock Squeeze-and-excitation networks.
\newblock In {\em Proceedings of the IEEE conference on computer vision and
  pattern recognition}, pages 7132--7141, 2018.

\bibitem{bradski2008learning}
Gary Bradski and Adrian Kaehler.
\newblock {\em Learning OpenCV: Computer vision with the OpenCV library}.
\newblock " O'Reilly Media, Inc.", 2008.

\bibitem{shrivastava2016training}
Abhinav Shrivastava, Abhinav Gupta, and Ross Girshick.
\newblock Training region-based object detectors with online hard example
  mining.
\newblock In {\em Proceedings of the IEEE conference on computer vision and
  pattern recognition}, pages 761--769, 2016.

\bibitem{ch2017total}
Chee~Kheng Ch'ng and Chee~Seng Chan.
\newblock Total-text: A comprehensive dataset for scene text detection and
  recognition.
\newblock In {\em 2017 14th IAPR International Conference on Document Analysis
  and Recognition (ICDAR)}, volume~1, pages 935--942. IEEE, 2017.

\bibitem{yuliang2017detecting}
Liu Yuliang, Jin Lianwen, Zhang Shuaitao, and Zhang Sheng.
\newblock Detecting curve text in the wild: New dataset and new solution.
\newblock {\em arXiv preprint arXiv:1712.02170}, 2017.

\bibitem{gupta2016synthetic}
Ankush Gupta, Andrea Vedaldi, and Andrew Zisserman.
\newblock Synthetic data for text localisation in natural images.
\newblock In {\em Proceedings of the IEEE Conference on Computer Vision and
  Pattern Recognition}, pages 2315--2324, 2016.

\bibitem{xu2019textfield}
Yongchao Xu, Yukang Wang, Wei Zhou, Yongpan Wang, Zhibo Yang, and Xiang Bai.
\newblock Textfield: Learning a deep direction field for irregular scene text
  detection.
\newblock {\em IEEE Transactions on Image Processing}, 2019.

\bibitem{tian2016detecting}
Zhi Tian, Weilin Huang, Tong He, Pan He, and Yu~Qiao.
\newblock Detecting text in natural image with connectionist text proposal
  network.
\newblock In {\em European conference on computer vision}, pages 56--72.
  Springer, 2016.

\bibitem{zhu2018sliding}
Yixing Zhu and Jun Du.
\newblock Sliding line point regression for shape robust scene text detection.
\newblock In {\em 2018 24th International Conference on Pattern Recognition
  (ICPR)}, pages 3735--3740. IEEE, 2018.

\bibitem{shi2017detecting}
Baoguang Shi, Xiang Bai, and Serge Belongie.
\newblock Detecting oriented text in natural images by linking segments.
\newblock In {\em Proceedings of the IEEE Conference on Computer Vision and
  Pattern Recognition}, pages 2550--2558, 2017.

\bibitem{li2018shape}
Xiang Li, Wenhai Wang, Wenbo Hou, Ruo-Ze Liu, Tong Lu, and Jian Yang.
\newblock Shape robust text detection with progressive scale expansion network.
\newblock {\em arXiv preprint arXiv:1806.02559}, 2018.

\end{thebibliography}

\end{document}